\pdfoutput=1
\documentclass{article}

\usepackage[final]{nips_2016}
\usepackage{authblk}
\usepackage[utf8]{inputenc} 
\usepackage[T1]{fontenc}    
\usepackage{hyperref}       
\usepackage{url}            
\usepackage{booktabs}       
\usepackage{amsfonts}       
\usepackage{nicefrac}       
\usepackage{microtype}      
\usepackage{amsmath}
\usepackage{graphicx}
\usepackage{color}

\title{Neural Symbolic Machines: \\ 
 Learning Semantic Parsers on Freebase \\
 with Weak Supervision (Short Version) \thanks{The long version with more details can be found at \url{https://arxiv.org/abs/1611.00020}.} }
\author{ 

  \bf{Chen Liang}\thanks{Work done while the author was interning at Google},
  \bf{Jonathan Berant}\thanks{Work done while the author was a visiting scholar at Google},
  \bf{Quoc Le},
  \bf{Kenneth Forbus},
  \bf{Ni Lao} \\
Northwestern University, Evanston, IL \\
\{chenliang2013,forbus\}@u.northwestern.edu \\
Tel-Aviv University, Tel Aviv-Yafo, Israel \\
joberant@cs.tau.ac.il \\
Google Inc., Mountain View, CA \\
\{qvl,nlao\}@google.com
}

\begin{document}

\maketitle

\begin{abstract}
Extending the success of deep neural networks to natural language understanding and symbolic reasoning requires complex operations and external memory. 
Recent neural program induction approaches have attempted to address this problem, but are typically limited to differentiable  memory, and consequently cannot scale beyond small synthetic tasks.
In this work, we propose the Manager-Programmer-Computer framework, which integrates neural networks with \textit{non-differentiable} memory to support \textit{abstract}, \textit{scalable} and \textit{precise} operations through a friendly \textit{neural computer interface}. 
Specifically, we introduce a Neural Symbolic Machine, which contains a sequence-to-sequence neural "programmer", 
and a non-differentiable "computer" that is a Lisp interpreter with code assist. 
To successfully apply REINFORCE for training, we augment it with approximate gold programs found by an iterative maximum likelihood training process.
NSM is able to learn a semantic parser from weak supervision over a large knowledge base. It achieves new state-of-the-art performance on \textsc{WebQuestionsSP}, a challenging semantic parsing dataset. Compared to previous approaches, NSM is end-to-end, therefore does not rely on feature engineering or domain specific knowledge.

\end{abstract}

\section{Introduction}

Deep neural networks have achieved impressive performance in classification and structured prediction tasks with full supervision such as speech recognition \cite{hinton2012deep} and machine translation \cite{sutskever2014sequence,bahdanau2014align,wu2016gnmt}. Extending the success to natural language understanding and symbolic reasoning requires the ability to perform complex operations and make use of an external memory. 
There were several recent attempts to address this problem in neural program induction \cite{graves2014neural,Neelakantan2015NeuralPI,reed2015neural,kaiser2015neural,zaremba2015reinforcement,graves2016ntm,andreas2016compose}, which learn programs by using a neural sequence model to control a computation component. 
However, the memories in these models are either low-level (such as in Neural Turing machines\cite{zaremba2015reinforcement}), or differentiable so that they can be trained by backpropagation. 
This makes it difficult to utilize efficient discrete memory in a traditional computer, and limits their application to small synthetic tasks.

To better utilize efficient memory and operations, we propose a Manager-Programmer-Computer (MPC) framework for neural program induction, which integrates three components:
\begin{enumerate}
\item A \textbf{"manager"} that provides weak supervision through input and a reward signal indicating how well a task is performed. Unlike full supervision, this weak supervision is much easier to obtain at large scale (see an example task in Section \ref{sec:webquestions}).
\item A \textbf{"programmer"} that takes natural language as input and generates a program that is a sequence of tokens. 
The programmer learns from the reward signal and must overcome the hard search problem of finding good programs. (Section \ref{sec:model}).
\item A \textbf{"computer"} that executes the program. It can use all the operations that can be implemented as a function in a high level programming language like Lisp. The \textit{non-differentiable} memory enables \textit{abstract}, \textit{scalable} and \textit{precise} operations, but it requires reinforcement learning. It also provides a friendly \textit{neural computer interface} to help the "programmer" reduce the search space by detecting and eliminating invalid choices (Section \ref{sec:language}).
\end{enumerate}

Within the MPC framework, we introduce the Neural Symbolic Machine (NSM) and apply it to semantic parsing. NSM contains a sequence-to-sequence neural network model ("programmer") augmented with a key-variable memory to save and reuse intermediate results for compositionality, and a non-differentiable Lisp interpreter ("computer") that executes programs against a large knowledge base. As code assist, the "computer" also helps reduce the search space by checking for syntax and semantic errors.
Compared to existing neural program induction approaches, the efficient memory and friendly interface of the "computer" greatly reduce the burden of the "programmer" and enable the model to perform competitively on real applications. On the challenging semantic parsing dataset \textsc{WebQuestionsSP} \cite{yih2016webquestionssp}, NSM achieves new state-of-the-art results with weak supervision. Compared to previous work, it is end-to-end, therefore does not require any feature engineering or domain-specific knowledge. 

\section{Neural Symbolic Machines}

Now we describe in details a Neural Symbolic Machine that falls into the MPC framework, and how it is applied to learn semantic parsing from weak supervision.

Semantic parsing is defined as follows:
 given a knowledge base (KB) $\mathbb{K}$, and a question $q=(w_1, w_2, ..., w_k)$, produce a program or logical form $z$ that when executed against $\mathbb{K}$ generates the right answer $y$.
Let $\mathcal{E}$ denote a set of entities (e.g.,  \textsc{AbeLincoln})\footnote{We  also consider numbers (e.g., ``1.33'') and date-times (e.g., ``1999-1-1'') as entities.}, and let $\mathcal{P}$ denote a set of properties (or relations, e.g., \textsc{PlaceOfBirthOf}). A knowledge base $\mathbb{K}$ is a set of assertions $(e_1,p,e_2) \in \mathcal{E} \times \mathcal{P} \times \mathcal{E}$, such as  (\textsc{Hodgenville}, \textsc{PlaceOfBirthOf}, \textsc{AbeLincoln})). 

\subsection{"Computer": Lisp interpreter with code assist}
\label{sec:language}

Operations learned by current neural network models with differentiable memory, such as addition or sorting, do not generalize perfectly to inputs that are larger than previously observed ones \cite{graves2014neural,reed2015neural}. 
In contrast, operations implemented in ordinary programming language are \textit{abstract}, \textit{scalable}, and \textit{precise}, because no matter how large the input is or whether it has been seen or not, they will be processed precisely. 
Based on this observation, we implement all the operations necessary for semantic parsing with ordinary non-differentiable memory, and allow the "programmer" to use them with a high level general purpose programming language.

We adopt a Lisp interpreter with predefined functions listed in \ref{tab-functions} as the "computer". The programs that can be executed by it are equivalent to the limited subset of $\lambda$-calculus in \cite{yih2015semantic}, but easier for a sequence-to-sequence model to generate given Lisp's simple syntax. Because Lisp is a general-purpose and high level language, it is easy to extend the model with more operations, which can be implemented as new functions, and complex constructs like control flows and loops.

A program $C$ is a list of expressions $(c_1 ... c_N)$. 
Each expression is either a special token "RETURN" indicating the end of the program,
or a list of tokens enclosed by parentheses "( $F$ $A_0$ ... $A_K$ )". $F$ is one of the functions in Table \ref{tab-functions}, which take as input a list of arguments of specific types, and, when executed, returns the denotation of this expression in $\mathbb{K}$, which is typically a list of entities, and saves it in a new variable. 
$A_k$ is $F$'s $k$th argument, which can be either a relation $p \in \mathcal{P}$ or a variable $v$. The variables hold the results from previous computations, which can be either a list of entities from executing an expression or an entity resolved from the natural language input.

\begin{table}[h!]
\centering
  \begin{tabular}{ cl}
  	\hline
    $($ \textit{Hop} $v$ $p$ $)$ $\Rightarrow$ $\{e_2 | e_1 \in v, (e_1, p, e_2) \in \mathbb{K}\}  $\\ 
    $($ \textit{ArgMax}  $v$ $p$ $)$ $\Rightarrow$ $\{e_1 | e_1 \in v, \exists e_2 \in \mathcal{E}: (e_1, p, e_2) \in \mathbb{K}, \forall e : (e_1, p, e) \in \mathbb{K}, e_2 \geq e \}$\\ 
    $($ \textit{ArgMin}  $v$ $p$ $)$ $\Rightarrow$ $\{e_1 | e_1 \in v, \exists e_2 \in \mathcal{E}: (e_1, p, e_2) \in \mathbb{K}, \forall e : (e_1, p, e) \in \mathbb{K}, e_2 \leq e \}$\\ 
    $($ \textit{Equal}  $v_1$ $v_2$ $p$  $)$ $\Rightarrow$ $\{e_1 | e_1 \in v_1, \exists e_2 \in v_2:  (e_1, p, e_2) \in \mathbb{K}\}  $ \\ 
    \hline
  \end{tabular}
\caption{Predefined functions. $v$ represents a variable. $p$ represents a relation in Freebase.  $\geq$ and $\leq$ are defined on numbers and datetime. 
}
\label{tab-functions}
\end{table}

To create a better \textit{neural computer interface}, the interpreter provides code assist by producing a list of valid tokens for the "programmer" to pick from at each step. 
First, a valid token should not cause a syntax error, which is usually checked by modern compilers. 
For example, if the previous token is "$($", the next token  must be a function, and if the previous token is "\textit{Hop}", the next token must be a variable. 
More importantly, a valid token should not cause a semantic error or run-time error, which can be detected by the interpreter using the value or denotation of previous expressions. For example, given that the previously generated tokens are "$($", "\textit{Hop}", "$v$", the next available token is restricted to the set of relations 
that are reachable from entities in $v$. By providing this \textit{neural computer interface}, the interpreter reduces the "programmer"'s search space by orders of magnitude, and enables weakly supervised learning on a large knowledge base. 

\subsection{"Programmer": key-variable memory augmented Seq2Seq model}
\label{sec:model}

The "computer" implements the operations (functions) and stores the values (intermediate results) in variables, which simplifies the task for the "programmer". The "programmer" only needs to map natural language into a program, which is a sequence of tokens that references operations and values in the "computer". 
We use a standard sequence-to-sequence model with attention and augment it with a key-variable memory to reference the values.

A typical sequence-to-sequence model consists of two RNNs, an encoder and a decoder. We used a 1-layer GRU \cite{cho2014:gru}, which is a simplified variant of LSTM, for both the encoder and the decoder. 
Given a sequence of words ${w_1, w_2 ... w_m}$, each word $w_t$ is mapped to a multi-dimensional embedding $q_t$ (see details about the embeddings in Section \ref{sec:experiment}). 
Then, the encoder reads in these embeddings and updates its hidden state step by step using:
$h_{t+1} = GRU(h_{t}, q_{t}, \theta_{Encoder})$, where $\theta_{Encoder}$ are the GRU parameters.
The decoder updates its hidden states $u_t$
by $u_{t+1} = GRU(u_{t}, c_{t-1}, \theta_{Decoder})$, where $c_{t-1}$ is the embedding of last step's output token $a_{t-1}$ (see details about the embeddings in Section \ref{sec:experiment}), and $\theta_{Decoder}$ are the GRU parameters.
The last hidden state of the encoder $h_{T}$ is used as the decoder's initial state. 
We adopt a dot-product attention similar to that of \cite{dong2016language}.
The tokens of the program ${a_1, a_2 ... a_n}$ are generated one by one using a softmax over the vocabulary of valid tokens for each step (Section \ref{sec:language}). 

\begin{figure}[h!]
\centering
\includegraphics[width=\textwidth]{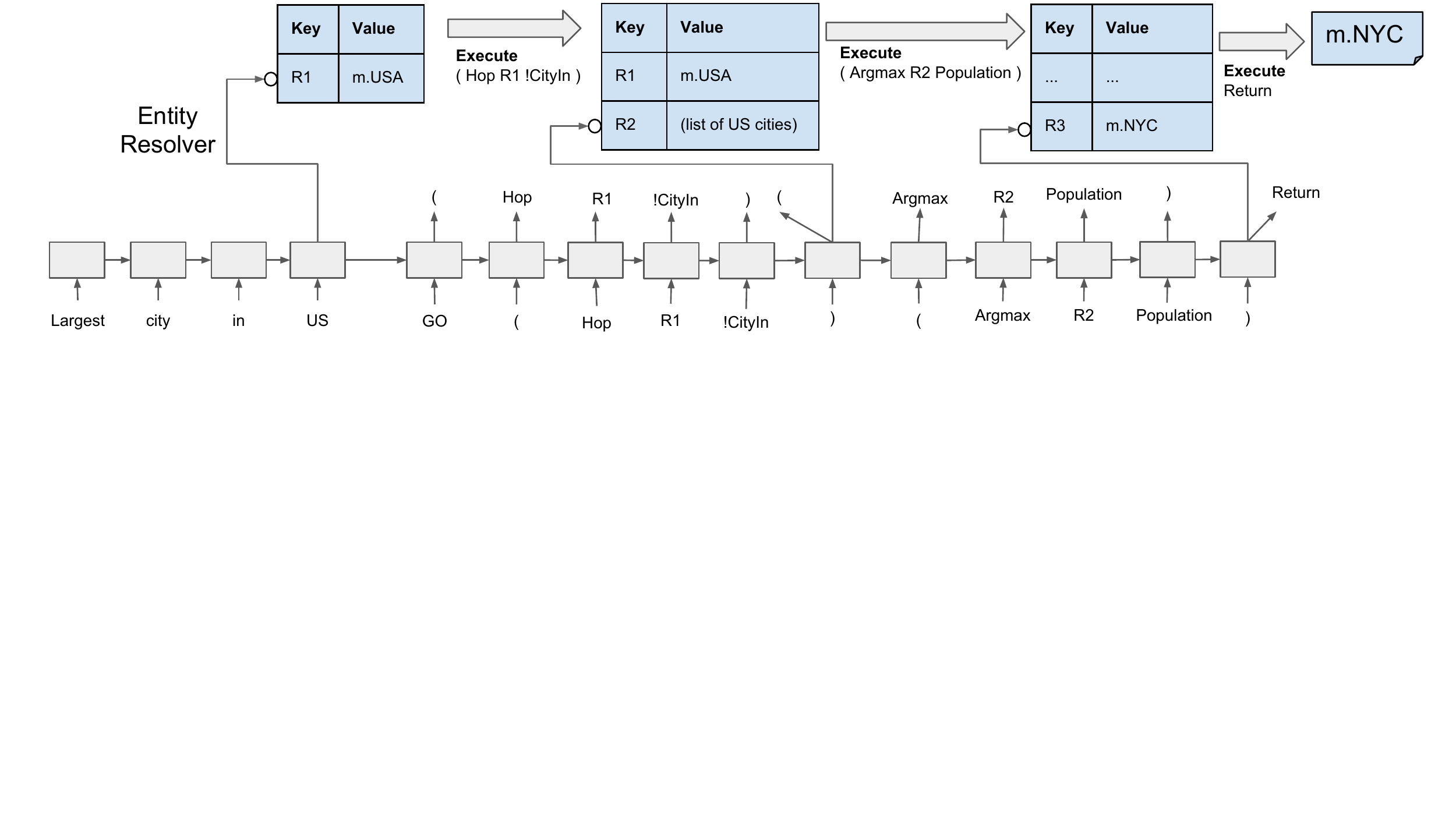}
\vspace{-2.0in}
\caption{\label{fig:parsing} Semantic Parsing with NSM. 
The key embeddings of the key-variable memory are the output of the sequence model at certain encoding or decoding steps.
For illustration purposes, we also show the values of the variables in parentheses, but the sequence model never sees these values, and only references them with the name of the variables such as ``R1''. A special token ``GO'' indicates the start of decoding, and ``RETURN'' indicates the end of decoding.}
\end{figure}
To achieve compositionality, we augment the model with a \textbf{key-variable memory} (Figure \ref{fig:parsing}). 
Each entry in the key-variable memory has two components:  a continuous multi-dimensional embedding key $v_i$, and a corresponding variable $R_i$ that references certain result in the "computer". 
During encoding if a token (\emph{"US"}) is the last token of a resolved entity (by an entity resolver), then the resolved entity id (\emph{m.USA}) is saved in a new variable in the "computer", and the key embedding for this variable is the average GRU output of the tokens spanned by this entity.
During decoding if an expression is completely finished (the decoder reads in "$)$"), it gets executed, and the result is stored as the value of a new variable in the "computer". This variable is keyed by the GRU output of that step.
Every time a new variable is pushed into the memory, the variable token is added to the vocabulary of the decoder.

\subsection{Training NSM with Weak Supervision} \label{sec:training}

To efficiently train NSM from weak supervision, we apply the REINFORCE algorithm \cite{Williams92simplestatistical,zaremba2015reinforcement}. However, the REINFORCE objective is known to be very hard to optimize starting from scratch. Therefore, we augment it with approximate gold programs found by an iterative maximum likelihood training process. 
During training, the model always puts a reasonable amount of probability on the best programs found so far, 
and anchoring the model to these high-reward programs greatly speeds up the training and helps to avoid local optimum. More details of the training procedure can be found in the long version. 

\section{Experiments and analysis} 
\label{sec:experiment}

\label{sec:webquestions}

Modern semantic parsers \cite{berant2014semantic}, which map natural language utterances to executable logical forms, have been successfully trained over large knowledge bases from weak supervision\cite{yih2015semantic}, but require substantial feature engineering. 
Recent attempts to train an end-to-end neural network for semantic parsing \cite{dong2016language,jia2016data} have either used strong supervision (full logical forms), or have employed synthetic datasets. 

We apply NSM to learn a semantic parser with weak supervision and no manual engineering. 
we used the challenging semantic parsing dataset \textsc{WebQuestionsSP} \cite{yih2016webquestionssp}, which consists of 3,098 question-answer pairs for training and 1,639 for testing. 
These questions were collected using Google Suggest API and the answers were originally obtained \cite{berant2013semantic} using Amazon Mechanical Turk and updated by annotators who are familiar with the design of Freebase \cite{yih2016webquestionssp}. We further separate out 620 questions in the training set as validation set.
For query pre-prosessing we used an in-house named entity linking system to find the entities in a question. The quality of the entity resolution is similar to that of \cite{yih2015semantic} with about $94\%$ of the gold root entities being included in the resolution results. 
Similar to \cite{dong2016language}, we also replaced named entity tokens with a special token "ENT". For example, the question \emph{"who plays meg in family guy"} is changed to \emph{"who plays ENT in ENT ENT"}. 

Following \cite{yih2015semantic} we use the last public available snapshot of Freebase KB. 
Since NSM training requires random access to Freebase during decoding, we preprocess Freebase by removing predicates that are not related to world knowledge (starting with "/common/", "/type/", "/freebase/")\footnote{Except that we kept ``/common/topic/notable\_types''.}, and removing all text valued predicates, which are rarely the answer.
This results in a graph with 23K relations, 82M nodes, and 417M edges.

We evaluate performance using the offical measures for \textsc{WebQuestionsSP}. Because the answer to a question can contain multiple entities or values, precision, recall and F1 are computed based on the output for each individual question. The average F1 score is reported as the main evaluation metric. The accuracy@1 measures the percentage of questions that are answered exactly. The comparison with previous state-of-the-art \cite{yih2016webquestionssp,yih2015semantic} is shown in Table \ref{tab-result}. Besides the better performance, our model does not rely on domain-specific rules or feature engineering.  
More ablation studies and analysis are included in the long version. 
\begin{table}[h!]
\centering
  \begin{tabular}{ l | r | r | r | r }
  	\hline
    Model & Avg. Prec. & Avg. Rec. & Avg. F1 & Acc.  \\ \hline \hline
    \textit{STAGG} & 67.3 & 73.1 & 66.8 & 58.8 \\ 
    \textit{NSM -- our model} & 70.9 & 74.8 & \textbf{69.0} & 59.4 \\
    \hline
    \hline
  \end{tabular}
\caption{Comparison to previous state-of-the-art, average F1 is the main evaluation metric. Our model achieves better results without hand-crafted rules and feature engineering.}
\label{tab-result}
\end{table}

\bibliographystyle{plain}
\bibliography{nips.bib}

\end{document}